\documentclass[a4paper,fleqn]{cas-dc}

\usepackage[numbers]{natbib}
\usepackage{lineno}
\modulolinenumbers[5]

\usepackage{hyperref}
\usepackage{url}
\usepackage{graphicx,subfigure,color}
\usepackage{amsthm}
\usepackage{amssymb}
\usepackage{amsmath,bm}
\theoremstyle{definition}
\newtheorem{myDef}{Definition}
\usepackage{algorithmic,algorithm}
\def\tsc#1{\csdef{#1}{\textsc{\lowercase{#1}}\xspace}}
\tsc{WGM}
\tsc{QE}
\tsc{EP}
\tsc{PMS}
\tsc{BEC}
\tsc{DE}
\begin{document}
%%%%%%%%%%%%%%%%%%%%%%%
\let\WriteBookmarks\relax
\def\floatpagepagefraction{1}
\def\textpagefraction{.001}

\shorttitle{A Concise Review of Recent Few-shot Meta-learning Methods}
\shortauthors{X. Li, Z. Sun, J.-H. Xue and Z. Ma}
\title[mode = title]{A Concise Review of Recent Few-shot Meta-learning Methods}

\author[lut]{Xiaoxu Li}

%% or include affiliations in footnotes:

\author[ucl]{Zhuo Sun}

\author[ucl]{Jing-Hao Xue}

\ead{}

\author[bupt]{Zhanyu Ma}

\address[lut]{School of Computer and Communication, Lanzhou University of Technology, China.}
\address[ucl]{Department of Statistical Science, University College London, U.K.}
\address[bupt]{Pattern Recognition and Intelligent System Laboratory, School of Artificial Intelligence, Beijing
University of Posts and Telecommunications, China.}

\begin{abstract}
Few-shot meta-learning has been recently reviving with expectations to mimic humanity's fast adaption to new concepts based on prior knowledge. In this short communication, we give a concise review on recent representative methods in few-shot meta-learning, which are categorized into four branches according to their technical characteristics. We conclude this review with some vital current challenges and future prospects in few-shot meta-learning.
\end{abstract}


\maketitle

\section{Introduction}

Deep learning has achieved a major breakthrough in large-scale image classification~\citep{krizhevsky2012ImageNet,simonyan2014very,szegedy2015going,gu2015recent, liu2017survey}. However, small-sample image classification such as few-shot learning is still a big challenge~\citep{santoro2016one,dong2018few,shu2018small,rahman2018unified}.
In this short communication, we present a concise review of recent representative meta-learning methods for few-shot image classification. We refer to such methods as {\it few-shot meta-learning} methods. After establishing necessary notation, we first mathematically formulate few-shot learning and offer a pseudo-coded algorithm for general few-shot training and evaluation. We then provide a taxonomy and a gentle review of recent few-shot meta-learning methods, to help researchers quickly grasp the state-of-the-art methods in this field. Finally we summarize some vital challenges to conclude this review with new prospects.

\section{The Framework of Few-shot Meta-learning}\label{framework}

\subsection{Notation and definitions}
We first establish the notation for few-shot learning. 

Suppose we have two datasets available: one base dataset $\mathcal{D}_{base}= \{(X_i,Y_i), Y_i \in \mathcal{C}_{base}\}_{i=1}^{N_{base}} $ and one novel dataset $\mathcal{D}_{novel} = \{ (\tilde{X}_j, \tilde{Y}_j ), \tilde{Y}_j \in \mathcal{C}_{novel} \}_{j=1}^{N_{novel}}$, where $(X_i, Y_i)$ is a tuple for the $i$th image with original feature vector $X_i$ and class label $Y_i$; $N_{base}$ and $N_{novel}$ denote the total numbers of observations in $\mathcal{D}_{base}$ and $\mathcal{D}_{novel}$, respectively; and the two class label sets $\mathcal{C}_{base} $ and $\mathcal{C}_{novel}$ are disjoint.
In few-shot learning, the task $\mathcal{T}=\mathcal{S} \cup \mathcal{Q}$ consists of a small set $\mathcal{S}$ of labeled support images and a set $\mathcal{Q}$ of query images from the same set of classes, such that a classifier $f$ is expected to correctly discriminate query images $\mathcal{Q}$ conditional on the small-size labeled support images $\mathcal{S}$. 
%Note that the labels are known for both $\mathcal{S}$ and $\mathcal{Q}$ during training, while only the labels of $\mathcal{S}$ are applied during evaluation.

In the cases of few-shot meta-learning, a meta-learner is trained to learn some prior or shared knowledge from $\mathcal{D}_{base}$, and then modified on tasks on $\mathcal{D}_{novel}$.  In this review, within a task $\mathcal{T}^{(k)}$, all support images are denoted by a set $\mathcal{S}^{(k)}$, similarly the query set by $\mathcal{Q}^{(k)}$, and images  should be from same set of classes $\mathcal{C}^{(k)}$, a subset of $\mathcal{C}$ where $\mathcal{C}$ can be $\mathcal{C}_{base}$, $\mathcal{C}_{novel}$ or their union $\mathcal{C}_{base} \cup \mathcal{C}_{novel}$. The process of generating tasks from $\mathcal{D}_{novel}$ is: $\{\tilde{\mathcal{T}}^{(k)}  = \tilde{\mathcal{S}}^{(k)} \cup \tilde{\mathcal{Q}}^{(k)} \}_{k=1}^{T_{novel}}$ are randomly sampled from the novel dataset $\mathcal{D}_{novel}$ by sampling the label sets $\{\tilde{\mathcal{C}}^{(k)} \}_{k=1}^{T_{novel}}$ from $\mathcal{C}_{novel}$ and subsequently sampling instances within those classes, where $T_{novel}$ is the total number of tasks we sampled on the novel dataset. A learner, after being trained on $\mathcal{D}_{base}$, is now required to learn to classify the query images of each task $\tilde{\mathcal{T}}^{(k)}$ after limited adaption via its small support set $\tilde{\mathcal{S}}^{(k)}$, for all $k$.
Since $\mathcal{C}_{base}$ and $\mathcal{C}_{novel}$ are disjoint, the tasks from $\mathcal{D}_{base}$ and $\mathcal{D}_{novel}$ are not directly related, but are linked via some transferable knowledge. Thus, a good learner should not only extract sufficient transferable knowledge among tasks but also fast adapt to novel tasks. Hence, in general, a few-shot meta-learning algorithm usually consists of two components, a meta-learner component and a task-specific learner component.

\theoremstyle{definition}

%\begin{myDef}
%(Small-sample classification) Suppose that the support set $\mathcal{S}$ is available for some task $\mathcal{T} = \mathcal{S} \cup \mathcal{Q}$ and $|\mathcal{S}|$ is small, this corresponds to small sample learning scenarios.
%Small-sample classification aims to learn a classifier $f$, conditional on $\mathcal{S}$ with or without some base dataset, to have excellent generalization performance that is evaluated on the query set $\mathcal{Q}$. Statistically, if averaging over the distribution of tasks, it is equivalent to minimizing true risk of the classifier $f$.
%\end{myDef}

Few-shot meta-learning is a typical way to achieve 
few-shot learning, which is a type of small-sample learning where the size $|\mathcal{S}|$ of support set $\mathcal{S}$ is small and the base dataset $\mathcal{D}_{base}$ is available. Hence we define them in turn.

\begin{myDef} (Small-sample learning)
A classifier $f$ is trained to learn some transferable prior knowledge from the base dataset $\mathcal{D}_{base}$, and then tuned on the support set $\tilde{\mathcal{S}}^{(k)}$, in order to correctly classify the query set $\tilde{\mathcal{Q}}^{(k)}$ of $\tilde{\mathcal{T}}^{(k)}$, for all $k \in \{1, 2, \ldots, T_{novel}\}$. This is equivalent to maximizing the generalization performance or minimizing the true risk of the classifier $f(\tilde{\mathcal{Q}}|\mathcal{D}_{base}, \tilde{\mathcal{S}})$, where $\tilde{\mathcal{S}}$ and $\tilde{\mathcal{Q}}$ are random variables as $\tilde{\mathcal{T}} = \tilde{\mathcal{S}} \cup \tilde{\mathcal{Q}}$ itself is a random sample from $\mathcal{D}_{novel}$.
\end{myDef}

\begin{myDef}
\label{def:few-shot learning}
(Few-shot learning) Let $\tilde{\mathcal{S}}_{c}^{(k)}$ be a subset of $\tilde{\mathcal{S}}^{(k)}$ that only contains images from the $c$th class, where the class $c$ belongs to the label set $\tilde{\mathcal{C}}^{(k)}$. If cardinality $|\tilde{\mathcal{S}}_{c}^{(k)}|$ is considerable small (e.g. from $1$ to $10$) for all $c \in \{1, 2, ..., |\tilde{\mathcal{C}}^{(k)}|\}$ and for all $k \in \{1, 2, \ldots, T_{novel}\}$, we refer to this as few-shot. In particular, if cardinality $|\tilde{\mathcal{S}}_{c}^{(k)}| = 1$ for all $c$ and all $k$, this is called one-shot classification; and when $|\tilde{\mathcal{S}}_{c}^{(k)}| = K$ and $|\tilde{\mathcal{C}}^{(k)}| = C$ for all $k$, this refers to $C$-way $K$-shot classification.
\end{myDef}

\begin{myDef}
(Few-shot meta-learning)
We refer to meta-learning algorithms specifically designed for few-shot classification as few-shot meta-learning. In general, such a meta-learning algorithm sets up a meta-learner component and a task-specific learner component, allowing information to flow among tasks and thus among base classes $\mathcal{C}_{base}$ and novel classes $\mathcal{C}_{novel}$, and  $\mathcal{D}_{base}$ is used to extract high-level knowledge rather than task-specific knowledge. 
\end{myDef}

The key of few-shot meta-learning is to extract and transfer knowledge from $\mathcal{D}_{base}$ to $\mathcal{C}_{novel}$. 
%Understanding learning conditional on the base dataset $\mathcal{D}_{base}$ is not sufficient to train a good classifier that fast adapts to novel concepts or classes. 
What to share, how to share and when to share are three components at the heart of few-shot meta-learning. For example, embedding layers are often shared in a rigid manner (e.g.~\cite{chen2019a}) in fine-tuning; parameters optimized on the base dataset can be regarded as a good initialization (e.g.~\cite{finn2017model, finn2018probabilistic}), for fast further learning conditional on few labeled samples from novel classes; and auxiliary information also helps few-shot learning, e.g.~attribute annotations related to images~\citep{Tokmakov_2019_ICCV}.

%In the framework of few-shot meta-learning, by constructing a meta-learner component at a global perspective, people can easily manipulate what and how and when the prior or transferable knowledge are learned or shared from the base dataset $\mathcal{D}_{base}$, and then expect to generalize to novel concepts or classes $\mathcal{C}_{novel}$ conditional on novel tasks.

\subsection{Training and evaluation of few-shot meta-learning}

Few-shot meta-learning models are usually trained and evaluated by forming few-shot episodes. An episode here is referred to a task $\mathcal{T}$. In addition to standard few-shot episodes defined by $C$-way $K$-shot, other episodes can also be used as long as they do not poison the evaluation in meta-validation or meta-testing, e.g.~incrementing query sets of tasks from novel classes with images from base classes~\citep{ren2018incremental}. In general, meta-training and meta-testing are implemented on $\mathcal{D}_{base}$ and $\mathcal{D}_{novel}$, and we have access to an extra dataset $\mathcal{D}_{val}$ with a set of classes $\mathcal{C}_{val}$ distinct to $\mathcal{C}_{base}$ and $\mathcal{C}_{novel}$ to evaluate model's performance and do model selection, e.g., choosing optimal number of epochs according to model's accuracy on tasks from $\mathcal{D}_{val}$. 

In this section, we give a general few-shot episodic training/evaluation guide in Algorithm~\ref{alg:general}, as an extension to the formulation in~\cite{sung2018learning}. 
For few-shot meta-learning, we can always design a deep neural network $f_{\theta}$ parametrized by $\theta$ as the classifier: we denote it as $f_{\theta}(\cdot |\mathcal{S}^{*}, \mathcal{D}_{base})$, where $\mathcal{S}^{*}$ is some support set.
%in general (i.e.~either $\in \mathcal{D}_{base}$ or $\in \mathcal{D}_{novel}$). 
To avoid redundancy, here we do not give exact forms in which meta-learning methods use the base dataset $\mathcal{D}_{base}$. For instance, fine-tuning and multiple phases of training are typical ways to reflect such conditional dependence. Note that in our notation $\mathcal{S}^{*}$ can also be a support set on the base classes $\mathcal{C}_{base}$ or even the whole base dataset $\mathcal{D}_{base}$, corresponding to the cases of meta-training or pre-training, respectively. The dependence on $\mathcal{S}^{*}$ and $\mathcal{D}_{base}$ is the hinge of few-shot meta-learning. Presenting such dependence is a purpose of this review. More details are discussed in Section \ref{sec:Methods}. 

In particular, when the following conditions in Algorithm~\ref{alg:general} are satisfied, it corresponds to standard $C$-way $K$-shot few-shot classification:
1) $\mathcal{D}_{base} \neq \emptyset$;
2) $|\mathcal{C}^{(e) *}| = C$ for all $e \in [E]$;
3) $|\mathcal{S}_{c}^{(e) *}| = K $ for all $e \in [E]$ and all $c \in [C]$;
4) $\mathcal{C}_{*}$ = $\mathcal{C}_{base}$ for training;
and 5) $\mathcal{C}_{*}$ = $\mathcal{C}_{novel}$ for evaluation. For few-shot classification problems, e.g., $C$-way $K$-shot classification, the performance of a learning algorithm is measured by its averaged accuracy on the query sets of the tasks generated on the novel dataset $\mathcal{D}_{novel}$ (i.e., the $15$th line of Algorithm~\ref{alg:general}).

\section{Methods of Few-shot Meta-learning}
\label{sec:Methods}

%With the development of deep learning in computer vision, recent works on small-sample image classification formalize it into reduced problems, e.g. $C$-way $K$-shot classification in few-shot learning. The main challenge of few-shot classification is deficiency of samples. The existing methods of few-shot classification mainly fall into four branches, \emph{meta-learning based methods, metric learning based method, data augmentation based methods,and regularization techniques for few-shot}. In this work, we give a selected review for meta-learning methods designed specifically for few-shot image classification.

\begin{algorithm}[htbp] 
\renewcommand{\algorithmicrequire}{\textbf{Input:}}
\renewcommand\algorithmicensure {\textbf{Steps:}}
\caption{General training/evaluation procedure of few-shot learning}  \label{alg:general} 
\begin{algorithmic}[1] 
\REQUIRE ~~\\
$\mathcal{D}_{base} = \{ \bm{X}_i, \bm{Y}_i; \bm{Y}_i \in \mathcal{C}_{base} \}_{i=1}^{N_{base}}$; $\mathcal{D}_{novel}=\{\tilde{\bm{X}}_{j},\tilde{\bm{Y}}_{j} ; \tilde{\bm{Y}}_{j} \in \mathcal{C}_{novel}\}_{j=1}^{N_{novel}}$; number of episodes $E$.\\
\ENSURE ~~\\
\STATE $\text{e} \leftarrow 0$. %\COMMENT{Initializing the loop.}
%\STATE $\text{A} \leftarrow 0$ \COMMENT{Initializing the mean accuracy.}
\REPEAT
\STATE $\text{e} \leftarrow \text{e}+1$
\STATE  Sample class label set $\mathcal{C}^{(e) *}$ from $\mathcal{C}_{*}$.
%\COMMENT{If $ \vert \mathcal{C}^{(e) *} \vert = |\mathcal{C}_{*}|$, and $\mathcal{D}_{base} = \emptyset$, this corresponds to non-based small-sample classification as we can set the number of training samples within each class to be small.}
\STATE  Sample ${M}^{(e) *}$ samples for each class within the set $\mathcal{C}^{(e) *}$ randomly, and randomly split into a support set $\mathcal{S}^{(e) *}$ and a query set $\mathcal{Q}^{(e) *}$.
\STATE  Compute the loss $L^{(e) *}$ on the query set $\mathcal{Q}^{(e) *}$ conditional on $\mathcal{S}^{(e) *}$ and $\mathcal{D}_{base}$ for the classifier $f_{\theta}(\bm{X}^{(e) *}_k | \mathcal{S}^{(e) *}, \mathcal{D}_{base})$ for all $\bm{X}^{(e) *}_k  \in \mathcal{Q}^{(e) *}$.
\STATE Record $\hat{\bm{Y}}^{(e) *}_{k}= f_{\theta}(\bm{X}^{(e) *}_k | \mathcal{S}^{(e) *}, \mathcal{D}_{base})$.
\IF{$Evaluation$}
\STATE $a^{(e)} = \frac{1}{|\mathcal{Q}^{(e) *}|} \sum_{k=1}^{|\mathcal{Q}^{(e) *}|} \delta(\bm{Y}^{(e) *}_{k} = \hat{\bm{Y}}^{(e) *}_{k})$.
\ELSIF{$Training$}
\STATE  Update all/part of model's parameters $\theta$ w.r.t. $L^{(e) *}$ using an optimizer. 
\STATE  $a^{(e)} = \frac{1}{|\mathcal{Q}^{(e) *}|} \sum_{k=1}^{|\mathcal{Q}^{(e) *}|} \delta(\bm{Y}^{(e) *}_{k} = \hat{\bm{Y}}^{(e) *}_{k})$.
\ENDIF
\UNTIL{$\text{e}=E$}.
\STATE \textbf{return} If $Training$, stop training $f_{\theta}$ and output mean accuracy $\frac{1}{E}\sum_{e=1}^{E}a^{(e)}$; if $Evaluation$, output mean accuracy.
\end{algorithmic}
\end{algorithm}

A few-shot meta-learning method aims to learn a task-specific network from a meta network designed for few-shot learning. Thus, the architecture of such a method usually contains two components, a meta-learner network and a task-specific learner network. The meta-learner component is to learn transferable prior knowledge from the base dataset $\mathcal{D}_{base}$. 
%While, the mechanism of extraction and transfer of knowledge can be designed to meet various assumptions and concepts. 
The existing few-shot meta-learning methods can be categorized into four branches according to their technical characteristics: {\it 1) learning an initialization, 2) generation of parameters, 3) learning an optimizer}, and {\it 4) memory-based methods}. We summarize the representative methods in each branch in Table~\ref{tab:the conclusion for literatures} and review them concisely and gently in the following four sections.

\begin{table*}[htbp]
\centering
\caption{Summary of few-shot meta-learning methods reviewed in this paper.}\label{tab:the conclusion for literatures}
\begin{tabular}{llll}
  \toprule
%\multicolumn{4}{l}{{\bf Few-Shot Meta-Learning for Image Classification}} \\ \hline
    Learning an Initialization    &  Generation of Parameters & Learning an Optimizer   &  Memory-based Methods  \\ \hline
    MAML~\citep{finn2017model} &   Learner \& Pupil Network~\citep{bertinetto2016learning}   &  Meta-Learner LSTM~\citep{ravi2017optimization}    & MANN-LRUA~\citep{santoro2016meta}   \\
    PLATIPUS~\citep{finn2018probabilistic} &    Meta-Network~\citep{munkhdalai2017meta} & LEO~\citep{rusu2019meta}     &  SNAIL~\citep{mishra2018simple}  \\
    TAML~\citep{Jamal_2019_CVPR}     &    LGM-Net~\citep{li2019lgm} &     &  CSN~\citep{munkhdalai2018rapid}  \\
    Baseline++~\citep{chen2019a} &    Dynamic FSL with Forgetting~\citep{gidaris2018dynamic} &     &    \\
    Compositional Image Rep.~\citep{Tokmakov_2019_ICCV}  &  wDAE-GNN~\citep{gidaris2019generating}  &     &    \\
    &   Weight Imprinting~\citep{qi2018low}   &     &    \\
    &    Incremental FSL with Attention~\citep{ren2018incremental}  &     &    \\ 
    & TAFE-Net~\citep{wang2019tafe}            &      & \\
    & MTL~\citep{sun2019meta} &  &\\
\bottomrule
\end{tabular}
\end{table*}

\subsection{Learning an initialization}

The first branch, learning an initialization, assumes that a shared initialization learned from the base dataset $\mathcal{D}_{base}$ can fast adapt to the unseen tasks from $\mathcal{D}_{novel}$. The underlying rationale is that the task-specific parameters are close to this shared global initialization for all the tasks generated from $\mathcal{D}_{base}$ and  $\mathcal{D}_{novel}$.
It can be interpreted and executed in the following two ways in recent few-shot meta-learning methods:
\begin{enumerate}
    \item To learn a global initialization conditional on the (giant) base dataset~\citep{finn2017model, finn2018probabilistic}. That is, algorithms can learn to learn by seeking a joint optimization on both support set and query set from tasks generated on the base dataset $\mathcal{D}_{base}$. Given such a task on $\mathcal{D}_{base}$, meta parameters are firstly adapted to the task-specific parameters \emph{w.r.t} the loss of support set (task adaption phase), and then the loss of the query set is applied to update the meta parameters (meta update phase). Such learning to learn ability is established from the episodic meta-training process on the base dataset $\mathcal{D}_{base}$, and then naturally the trained models would have similar ability on the tasks from the novel dataset $\mathcal{D}_{novel}$.
    \item To fine-tune the trained parameters on the base dataset $\mathcal{D}_{base}$ via conditioning from few labeled images on the novel dataset $\mathcal{D}_{novel}$, e.g.~\cite{chen2019a, Tokmakov_2019_ICCV}. 
\end{enumerate}

Model-agnostic meta-learning ({\it MAML}), proposed by~\citet{finn2017model}, falls into the first way of learning an initialization. That is, it seeks to find a globally optimal initialization of parameters. During the meta-training procedure of {\it MAML}, the algorithm seeks to update the task-specific parameters and the global initialization jointly in an iterative manner. More specifically, given the current value of the global initialization, {\it MAML} performs a certain number of stochastic gradient descent steps by using the loss on the support set of a specific task. The loss from applying the task-specific parameters to the query set is then used to update the global initialization. In order to take model's uncertainty into consideration, \citet{finn2018probabilistic} proposed a generalized version of {\it MAML} that learned the posterior distribution of parameters given the support set of a task. Then, by sampling from the posterior, they constructed a posterior predictive model for the query set of the task. The inference of posteriors is achieved by variational inference and re-formulating {\it MAML} as a graphical model that introduces conditional independence; and the meta-learning component and task-specific component are integrated in a single neural network.

\citet{Jamal_2019_CVPR} considered that a meta model trained on the base dataset (e.g.~{\it MAML}) could be biased towards some tasks, i.e.~having diverse levels of learning for different tasks, which potentially results in large variation in the performance on novel tasks. Thus, they proposed a novel task-agnostic meta-learning algorithm ({\it TAML}), which aims to train an initial model that is unbiased to all tasks. During the meta-training process, the task-agnostic property of {\it TAML} is established by either maximizing the entropy reduction for each task or minimizing the inequality in performance of various tasks. {\it TAML} achieves the state-of-the-art performance in $5$-way $1$-shot and $5$-way $5$-shot classification on the {\it Onimiglot} dataset.

\citet{chen2019a} demonstrated that the modified baseline methods (denoted by {\it Baseline++}) could also have comparable performance to the state-of-art methods on both the {\it Mini-ImageNet} and {\it CUB-200-2011} datasets. 
Either of these baseline and {\it Baseline++} neural networks can be decomposed into two parts: a convolution embedding network and a classifier network. 
A baseline method uses the second way: the fine-tune strategy, which firstly learns a giant classification problem for all classes on the base dataset $\mathcal{D}_{base}$, and then fine-tunes parts of those parameters on the novel dataset $\mathcal{D}_{novel}$. In~\cite{chen2019a}, at the fine-tuning stage, they only keep the learned embedding part and set up a new classifier fit for the $C$-way $K$-shot problems on tasks generated from $\mathcal{D}_{novel}$. The parameters of the new classifier are fine-tuned by stochastic gradient descent of the loss on the support images of a novel task, and then the whole network is used to predict the query images from the same task. 
The classifier network of a standard baseline method consists of a linear mapping layer and a softmax activation function. The standard baseline method only learns the parameters of the linear mapping at the fine-tuning stage; a modified baseline method replaces the linear mapping layer by a layer that computes cosine distance between each image's deep representation and the learned parameters of the linear mapping layer (learned at the fine-tuning stage).
\citet{chen2019a} offer an appealing new view on few-shot learning and generously make publicly available the source code and all model implementations in a fair evaluation setting.

Based on the baseline methods in~\cite{chen2019a}, \citet{Tokmakov_2019_ICCV} proposed to learn an image representation that could be decomposed into parts corresponding to attribute annotations. This is achieved by incorporating additional regularization terms that constrain either hard via distance or softly between the embeddings of an image and attribute annotations. Here, the soft constraint is a constraint between the embeddings of an image and attribute annotations modified by a part of the image encoding. The proposed neural network is firstly trained on the base dataset $\mathcal{D}_{base}$, and then fine-tuned with those additional regularization terms conditional on the labeled images from each novel task. 
%and then the whole network is used to predict the class labels for the unlabeled images from the same task.

\subsection{Generation of parameters}
The second branch focuses on rapid generation of parameters of task-specific neural networks from a meta-learner.

Back to 2016, \citet{bertinetto2016learning} proposed to use a meta-learner to predict the parameters of a {\it pupil network} for one-shot classification. Prior information is extracted from the base dataset to the meta-learner named {\it learnet}. {\it Pupil networks} serve as task-specific networks. The predicted parameters of a pupil network are generated by using a feed forward function without costly iterative optimization, thus providing fast computation. Interestingly, compared to probabilistic {\it MAML}~\citep{finn2018probabilistic}, it uses a learnable deterministic function to perform parameter generation. Although their network shares a lot in common with {\it Siamese Networks}~\citep{koch2015siamese}, a notable difference from {\it Siamese Networks} is that their feed forward mapping also changes the meta-learner by using the output of the mapping to parametrize some linear representation layers of the meta-learner. Thus, it is a dynamic procedure as the parameters of the meta-learner are no longer fixed. The meta-training process is end-to-end, but it samples in a slightly different way from Algorithm \ref{alg:general}. That is, for each training epoch, it samples thousands of triplets made up of a query image, a support image and an indicator that indicates whether they are from identical class.

Similarly, \citet{munkhdalai2017meta} proposed {\it Meta Networks} which consisted of a meta-learner and a base-learner for one-shot classification, of which the training procedure followed standard episodic meta-training. The meta-learner is used to perform fast parameter generation for itself and the base-learner by minimizing both the representation loss and task loss across various tasks with an attention mechanism. It outperforms {\it Siamese Networks}~\citep{koch2015siamese}, {\it MANN}~\citep{santoro2016meta} and {\it Matching Network}~\citep{vinyals2016matching} on the {\it Omniglot} dataset. The {\it LGM-Net} proposed by~\cite{li2019lgm} also employed a meta-learner and a base-learner, denoted by {\it MetaNet} and {\it TargetNet} (which was set to be {\it Matching Network}), at meta level and task level respectively, of which the training process also followed standard episodic meta-training. Fast parameter generation is achieved by {\it MetaNet} through learning the distribution of functional parameters of task-specific {\it Matching Networks} conditional on support sets of tasks.

Apart from above methods, \citet{gidaris2018dynamic} believed that few-shot learning algorithms should have fast adaption to novel classes $\mathcal{C}_{novel}$ without forgetting the base classes $\mathcal{C}_{base}$. This is achieved by combining an attention-based classification weight generator and a cosine-based convolution classifier which allows to learn both base and novel classes even at the testing stage. That is, the weight generator is set as the meta learner that takes both the deep features of a novel class and the trained classification weights of base classes as input, to generate classification weights for the associated novel class (and optionally the base classes). The cosine-based convolution network classifier\footnote{A standard convolution network classifier consists of a convolution embedding network and a classifier network, of which the last layer is a linear layer with weights $W$, which produces class membership by a softmax activation function.} further measures the cosine similarity between these weights and the deep features of a query image to give its probability scores for base and novel classes jointly. The training procedure of \citet{gidaris2018dynamic} is split into two stages, distinct from standard episodic training in~\cite{vinyals2016matching}. The first stage is to learn parameters of whole network excluding the weight generator on the base dataset $\mathcal{D}_{base}$, while the second stage is to train the weight generator by taking some classes from $\mathcal{C}_{base}$ as if they were novel. Based on the same belief, in~\cite{gidaris2019generating}, they further proposed to use denoising auto-encoders (as graph neural networks) to generate classification weights of novel classes and base classes jointly conditional on support sets of tasks from the novel dataset and the base dataset with episodic training processes following~\cite{vinyals2016matching}, which outperformed their former work in~\cite{gidaris2018dynamic} on {\it Mini-ImageNet}.

Rather than employing a meta-learner and a task-specific learner separately, \citet{qi2018low} proposed to imprint weights of novel classifiers by directly copying the normalized feature maps of novel training examples (using the average and then normalization if multiple training examples per class). Their contribution is to add normalization to the embeddings and weight matrix of the last layer in a standard convolution network classifier, by which the interpretation of inner-product, Euclidean distance and cosine distance are thus unified. The embedding network is learned on all base data $\mathcal{D}_{base}$ and then fine-tuned on novel classes $\mathcal{C}_{novel}$ to meet the uni-modality assumption for each class. Their weight imprinting method outperforms the nearest neighbors approach on the {\it CUB-200-2011} dataset.

Similarly, \citet{ren2018incremental} also wanted to solve the catastrophic forgetting on the base classes $\mathcal{C}_{base}$ in addition to fast adaption to novel classes $\mathcal{C}_{novel}$. Their main idea is to use an additional regularization on the weights ($W_{base}$) of base-class classifier, which is measured by an attention attractor network. The whole training includes two stages. A pre-training stage is to learn a good representation and so-called slow weights ($W_{base}$) of the top fully connected layer of the classifier. Then, an incremental few-shot episodic training is designed to increment novel classes into training via an episodic style. That is, modifying each task $\mathcal{T}$ in Algorithm \ref{alg:general} such that its support set $\mathcal{S}$ contains novel classes and that its query set $\mathcal{Q}$ have both novel and base classes. On the {\it Mini-ImageNet} dataset, the method outperformed those in~\cite{gidaris2018dynamic} and~\cite{qi2018low}.

While the above methods mainly focused on generating parameters of classifier blocks, \citet{wang2019tafe} proposed {\it TAFE-Net} that tuned task-specific feature embedding based on the generic embedding of a meta-learner. Their model consists of a meta-learner and a prediction network. The task-aware feature embedding is achieved by using the meta-learner to generate task-specific feature layers of the prediction network. Similarly, \citet{sun2019meta} proposed a meta-transfer learning method ({\it MTL}), which aimed to generate task-specific feature extractors mainly by a learnable scaling and shifting process on pre-trained feature embedding (i.e.~pre-training a feature extractor using all data points in $\mathcal{D}_{base}$) and had similar fine-tuning steps as those in~\cite{chen2019a}. They slightly modified the standard episodic meta-training (in Algorithm \ref{alg:general}) and proposed a novel meta-training scheme that put more focus on hard tasks by sampling extra samples from the classes that the learner was not good at. On {\it Mini-ImageNet}, the {\it MTL} achieves the state-of-the-art performance, for $5$-way $1$-shot classification.

\subsection{Learning an optimizer} 
The third branch, learning an optimizer, trains the meta-learner how to optimize all or part of model's parameters indirectly, e.g.~optimizing in a latent parameter space~\citep{rusu2019meta}.

Inspired by similar forms of updates for cell states in LSTMs\footnote{LSTM is the shorthand notation for Long Short Term Memory.} and in standard stochastic gradient-based optimizers (e.g.~ADAM, SGD), \citet{ravi2017optimization} proposed an LSTM-based meta-learner to learn the exact task-specific optimization of a classifier (i.e.~a convolution network classifier in their work) in the few-shot regime, and also learn good initialization values for the parameters of task-specific learner. Their main contribution is to represent parameter optimization of a task-specific classifier by the evolution of LSTM's cell states. 
Their work also uses a standard episodic meta training/evaluation as in Algorithm~\ref{alg:general}.

Considering that previous gradient-based meta-learning methods, e.g.~{\it MAML~\citep{finn2017model}}, used for few-shot learning have practical difficulties in optimization on high-dimensional parameter spaces, \citet{rusu2019meta} proposed the latent embedding optimization ({\it LEO}) that learned a data-dependent latent generative representation of model parameters and performed gradient-based meta-learning in this low-dimensional latent space. {\it LEO} has a similar learning algorithm to that of {\it MAML}, consisting of inner loops (for getting task-specific values given the current global initialization) and outer loops (for updating the global initialization). Their work also uses a standard episodic meta-training as in Algorithm~\ref{alg:general}. To instantiate data-dependent latent representation of model's parameters, images from the support set $\mathcal{S}$ of a task $\mathcal{T}$ pass through the combination of an encoder and a {\it Relation Network}~\citep{sung2018learning}. Their main contribution is to tailor the inner optimization loop of {\it MAML} such that task-specific parameters are learned from current global initialization by back-propagating loss on the support set $\mathcal{S}$ through the decoder within each task $\mathcal{T}$. Subsequently, similarly to {\it MAML}, the loss on the query set $\mathcal{Q}$ is used to update global initialization.
{\it LEO} achieves the state-of-the-art performance on the {\it Mini-ImageNet}.

\subsection{Memory-based methods}
Memory-based methods, the fourth branch, aim to solve few-shot meta-learning with memory resources. Generally, these methods are designed to access, write, read and use memory efficiently for few-shot classification problems.

For example, \citet{santoro2016meta} proposed to use memory-augmented neural networks for meta-learning that consisted of a controller (e.g.~LSTM) and an external memory module,  and came up with a novel content-based method to the excess external memory, referred to as {least recently used access} ({\it LRUA}). The proposed {\it LRUA} iteratively computes {\it usage}, {\it read}, {\it write} and {\it least-used} weights, and writes to external memories based on those {\it write} weights such that only zeroed memory slots and previously used slots are accessed. Within a task $\mathcal{T}$, images $\{x_t\}_t$ and corresponding responses $\{y_t\}_t$ enter as a temporal sequence such that the label of image $x_t$ is only available at time $t+1$. Therefore, the model is required to predict the response of $x_t$ given all past labeled images at time $t$. However, in terms of few-shot image classification problems, the proposed method was only tested on the {\it Omniglot} dataset. \citet{mishra2018simple} proposed a class of generic meta-learner architectures, called simple neural attentive learner ({\it SNAIL}), that combined temporal convolutions and soft attention for leveraging information from past episodes and for pinpointing specific pieces of information, respectively. Compared with {\it Meta Networks}~\citep{munkhdalai2017meta}, {\it SNAIL} obtains better performance on {\it Mini-ImageNet} and {\it Omniglot}.

\citet{munkhdalai2018rapid} proposed a neural mechanism, conditional shifted neurons ({\it CSNs}), which was capable of extracting conditioning information and producing conditional shifts for prediction in the process of meta-learning, and could be further incorporated into CNNs and RNNs. Their model also contains a meta-learner and a task-specific learner which receives conditional shifts from the meta-learner. The meta-learner extracts and uses conditional information (e.g.~error gradient information) to generate memory values for images within a task $\mathcal{T}$ at description time, and generates query keys of a query image by a key function in order to obtain the value of conditional shift. 
Under the implementation of LSTM and ResNet~\citep{He_2016_CVPR} backbone with {\it CSNs}, their work achieved better performance on the {\it Mini-ImageNet} and {\it Omniglot} datasets compared to widely-used neural network architectures, namely {\it adaCNN}, {\it adaResNet} and {\it adaLSTM}.

\section{Some Remaining Challenges}\label{trends}
Along with the promising performance of the few-shot meta-learning, there still remains some vital challenges, as well as irresistible trends.

The main challenge of few-shot learning is the deficiency of samples. The current few-shot meta-leaning methods try to solve this problem by extracting transferable or shared knowledge, e.g., a global initialization of parameters, from an auxiliary dataset through meta-training. Even though these knowledge can be learned, it is still difficult to train a model from few labeled samples. Parameter-generation based methods solve this problem by directly generating the parameters of the task-specific learner to mitigate the difficulty of training on novel data. However, these methods lack theoretical and in-depth analyses for choosing specific forms of parameter-generation. There is still a need for obtaining few-shot meta-learning algorithms with good generalization ability conditional on few labeled samples. How to construct better meta-learners, more effective task-specific learners, cross-domain few-shot meta-learners,  as well as multi-domain few-shot meta-learners will draw more attention in the future.

\paragraph{A better and more diversified meta-learner.}  A meta-learner can provide knowledge to a task-specific learner to mitigate the deficiency of samples. 
%However, when a meta-learner is over-trained or under-trained on a series of tasks, the knowledge learned by \textcolor{red}{the} meta-learner \textcolor{red}{can} be invalid for a new task~\citep{Jamal_2019_CVPR}. 
However, when a few-shot meta-learning algorithm has uneven performance on a series of tasks, the knowledge learned by the meta-learner can lead to large uncertainty in performance for novel tasks from unseen classes~\citep{Jamal_2019_CVPR}. Therefore, developing meta-learning algorithms with appropriate fitting ability is still full of challenges for few-shot learning. Apart from the existing  few-shot meta-learning methods, meta-learning methods with diversified emphases, such as learning a suitable loss function or learning a network structure, will also be valuable to explore.

\paragraph{A more effective task-specific learner.} 
An effective task-specific learner needs to make good use of the knowledge transferred from the meta-learner. Since the existing deep feature extractors are not tailored for few-shot learning, it remains vital for us to develop a feature extractor for task-specific learners that learns more discriminative features from only one or few labeled images. In addition, many loss functions, including some large-margin loss functions, are also not specifically designed for few-shot scenarios. Thus, it is important that task-specific learners are built on the loss functions that can ensure the robustness and performance of models.

\paragraph{Cross-domain few-shot meta-learning.} In practice, $\mathcal{D}_{base}$ and $\mathcal{D}_{novel}$ can be from different domains; such classification problems demand cross-domain few-shot learners. Experiments in the existing studies have shown that if the novel dataset $\mathcal{D}_{novel}$ is from a quite different domain, most of few-shot meta-learning methods fail to perform well on the novel tasks generated from $\mathcal{D}_{novel}$, because they usually assume that $\mathcal{D}_{base}$ and $\mathcal{D}_{novel}$ are from the same domain. Up to present, there are only few domain adaptation proposals for few-shot image classification. Therefore, it merits further exploration on cross-domain few-shot meta-learning.

\paragraph{Multi-domain few-shot meta-learning.} Furthermore, existing few-shot meta-learning methods usually assume that $\mathcal{D}_{base}$ is from a single domain.
If a meta-leaner can learn transferable knowledge from $\mathcal{D}_{base}$ consisting of multi-domain data, the meta-learner will be expected to have better generalization ability.
It is also easy in reality to construct a base dataset $\mathcal{D}_{base}$ from different domains.
Hence, multi-domain few-shot meta-learning is another topic worthy of research.

\bibliographystyle{cas-model2-names}
\bibliography{conciseref}

\end{document}